# A Finite State Transducer Based Morphological Analyzer of Maithili Language

Raza Rahi, Sumant Pushp[*], Arif Khan, Smriti Kumar Sinha
Department of Computer Science and Engineering
Tezpur University, Assam, India
[*]{sumantpushp@gmail.com}

**Abstract.** Morphological analyzers are the essential milestones for many linguistic applications like; machine translation, word sense disambiguation, spells checkers, and search engines etc. Therefore, development of an effective morphological analyzer has a greater impact on the computational recognition of a language. In this paper, we present a finite state transducer based inflectional morphological analyzer for a resource poor language of India, known as Maithili. Maithili is an eastern Indo-Aryan language spoken in the eastern and northern regions of Bihar in India and the southeastern plains, known as *tarai* of Nepal. This work can be recognized as the first work towards the computational development of Maithili which may attract researchers around the country to up-rise the language to establish in computational world.

## 1 Introduction

Morphology is the study of structure and formation of words by focusing on their internal structure. It is mainly concerned with analyzing individual word into their component and studied in two different aspects: inflectional morphology and derivational morphology [9]. They differ in the context that, when word stem of a particular class is combined with a morpheme, then first one result in a new word belongs to same class and later one result in a new word belongs to different class, where class means any grammatical category and morpheme refers to the smallest meaning bearing unit that can't be further divided into smaller meaningful parts.

Maithili is a highly inflectional language; it has the capability of generating hundreds of new words from a single root. Therefore, to establish the foundation of application like POS tagger, Machine Translator and many more others, development of morphological analyzer is essential for a language of having glorious literature. Morphological analyzer for many Indian languages have been reported in the literature in past [10,11,12], but works on Maithili do not exist as to best of our knowledge.

Number of research [13,14,15] have been done for developing morphological analyzers in various languages based on finite-state transducers. Beesley [13] showed how the morphotactics and the variation rules of Arabic have been described using only finite state operation and implemented a significant morphological analyzer using this approach. Beesley [14] designed a morphological analyser and generator for Aymara language using Xerox finite-state machine. Megerdoomian [15] described some of the challenges encountered in a computational morphological analysis of Persian and developed a morphological analyzer for Persian using Xerox finite-state tools.

In Indian language context, Kumar et al. [16] developed a morphological analyzer for Hindi. They used SFST tool for generating the finite states. A morphological generator was developed for Kannada language using FST [17]. Sahoo [18] described a finite state morphological analyzer for the nouns in Oriya. He specified the co-occurrence restrictions of the morphemes in a nominal form and used the FSA to determine whether an input string of morphemes makes up an Oriya noun or not. Bapat et al. [19], reported a paradigm-based finite state morphological analyzer for Marathi language. Vishal et al. [10] presents the morphological analysis and generator tool for Hindi Language using paradigm approach for windows platform having GUI. Nayan et al. [20] used the finite state approach to analyze morphology of the Bishnupriya Manipuri language.

In the era of computational environment, a language is recourse poor if it has less computational support, and Maithili suffers from the same. Therefore, to encourage the possibility of research we have designed a tool to analyze the morphological fragments of Maithili language. Xerox has built in Finite State Transducer to perform morphological analysis for many languages like English, French etc. There are numbers of tool available for the construction of FST based Morphological Analyzer among which XFST (Xerox Finite State Transducer), SFST, OFST are popular. The work presented in the paper uses the XFST tool for generating FST.

The construction of morphological analyzer is divided into two phases. In the first phase we build the lexicon file with the help of lexicon generator. The whole process of lexicon generation is done by using the raw corpus. Unique words from the corpus are extracted and sorted to make the task of processing of the words manually easier. These words are manually classified into different classes and according to their inflectional types. The Morphological analyzer needs a dictionary of root words, file containing FST rules and the dictionary of suffixes where the inflectional rules are hand written.

The organization of the paper is as follows; Section 2 describes the background and prerequisite required before the development of an analyzer. Section 3 and section 4 shows the proposed methodology and result of experiment respectively. Finally, section 5 concludes the work with future directions.

## 2 Background and Prerequisite

Maithili is an Eastern Indo-Aryan language spoken by a total of about 34 million people of northern India and Nepal of which 2.8 million are residents of Nepal [1,3]. In the past, Maithili was regarded as dialect of Bengali or Eastern Hindi or Bihari however, later it is recognized as a distinct language [4,5,6,8]. The first Grammar of Maithili was written by George Abraham Grierson in 1881 and entitled 'An Introduction to the Maithili language of North Bihar' [2]. The Grammatical studies on Maithili are roughly divided into five categories (a) Traditional (b) Historical (c) Pedagogical (d) Structural/descriptive, and (e) Lexicographical (1). Like many Indian languages, Maithili is a highly inflectional language and having the scope of generating many words from a single Noun, Adjective or Verb [7,21].

The major milestone required to be achieved before the development of an analyzer was to understand the pattern of inflection. We have analyzed the word structure of the Maithili language with the help of Maithili dictionary and found an exhaustive amount of pattern classified on the basis of parts of speech. For example, the suffix which are added to masculine nouns forms in order to form feminine are -*in, -ni, and –ain* represented in table 1.

**Table 1.** Noun inflection in Maithili for gender.

| Masculine | | Feminine |
|---|---|---|
| बाघ (*Bagh*) | 'tiger' | बाघिन (*Baghin*) |
| जात (*Jat*) | 'a caste name' | जातिन (*Jatin*) |
| दास (*Das*) | 'servant/slave' | दासिन (*dasin*) |

Many such forms of inflections have analysed before the development of an automated morphological analyzer. For example, the noun has two genders-masculine and feminine, words derived direct from the Sanskrit, which were originally neuter, generally become masculine in Maithili; Whereas Maithili pronouns are marked for person, number and case but not for gender. Adjectives in Maithili show no number or case distinctions. Gender distinctions are shown, but only marginally. Similarly, in the Maithili verb system there are no distinctions of number (i.e. singular and plural) and gender.

## 3 Proposed Methodology

The presented morphological analyzer works in two phases by generating lexicon and producing morphological detail using Xerox tool. Xerox combined both the Morphotactics and the orthographic rules into single lexical transducer [6]. It also supports UTF-8 characters coding which is very important for the Implementation of Maithili computational morphology.

### 3.1 Lexicon Generation

The lexicon of a language is its vocabulary. It is an explicit list of every word of the language. It is cumbersome to list every word in a language. Hence generally computational lexicons are used for this purpose. The corpus from LDC-IL(linguistic Data Consortium for Indian Language) has been used to generate the lexicon of Maithili. Thereafter unique words from the corpus are extracted and sorted to make the task of processing of the words manually easier. These words are then manually classified into various classes and inflection types.

A lexical description is always typed into a text file which consists of an optional *Multichar_Symbols* declaration, followed by an optional Declarations section, followed by one or more named LEXICONS, of which exactly one should be named LEXICON Root.

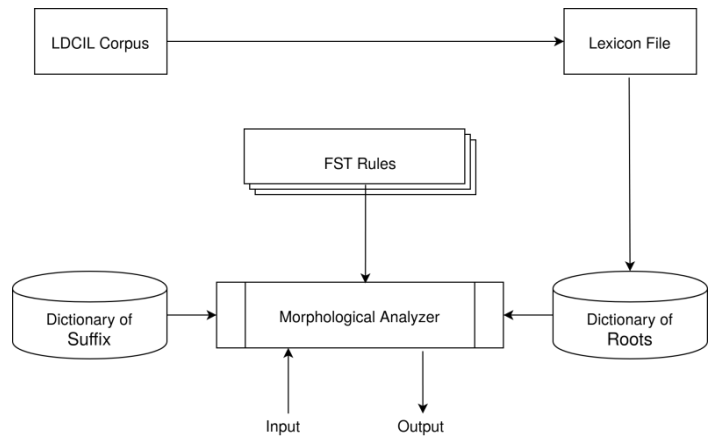

**Fig. 1**. System Architecture for Maithili Morphological Analyzer

### 3.1 Morphology Generation

Morphological analyzer takes input from dictionary of suffix, dictionary of root, and FST rules, where FST rules are handwritten and coded with XFST tool. Dictionary of root consists of all existing root of Maithili belongs to LDC-IL corpus. The Dictionary of Suffix contains a list of suffixes with their morpho-syntactic feature values like gender, number, person and other relevant morphological information stored in the form of a dual field list, which deals only with inflectional suffixes not derivational.

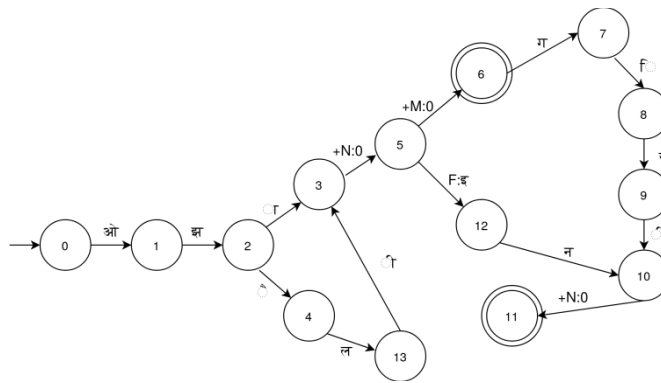

**Fig. 2**. Diagrammatic Representation of Example Words

Figure 2 represents a state network is a finite state transducer representing the morphological analyzer for Maithili language. The network illustrates the transitions for four words; ओझा, ओझागिरी, ओझाइन, ओझैली. Single circles represent a state and a double circle represents a final state(s). The numbers inside the circle is the state number, and the circle with a value 0 is the start state of the FST. The morphological analyzer starts at the initial state and goes through a sequence of states by computing the morphemes. If it matches the symbol on an arc leaving the present state, then it moves to the next state through that arc and moves to the next symbol of the input word and successfully recognizes all the morphemes in an input string. Table 2 shows the output of given words.

**Table 2.** Output of given example words

| Input Word | Output |
|---|---|
| ओझा | ओझा + N + M |
| ओझागिरी | ओझागिरी + N + M |
| ओझाइन | ओझा + N + F |
| ओझैली | ओझैली + N + M |

## 4 Experiments and Results

We tested Maithili morphological analyzer on LDC-IL corpus of around 855,430 words for performance evaluation and found no instance of failure on inflectional form as long as root belongs to lexicon file. Table 2, 3, 4 shows various results generated by Maithili morphological analyzer in the format of; *Root(suffix) + POS + Number + Gender*.

### 4.1 Experiment 1

Outcome of the first experiment was verified manually to test 16,572 Maithili noun extracted from the corpus. The system has correctly identified the root and provided the morphological analysis for 15,246 nouns. Table 2 shows the correct inflection type and output for few noun extracted by the analyzer.

**Table 3.** Result for Noun Inflection.

| Input Words | Inflection Type | Output |
|---|---|---|
| ओझा | Gender Inflection | ओझा + Noun + SG + Masculine |
| ओझाइन | | ओझा(आइन) + Noun + SG + Feminine |
| मास्टर | Gender Inflection | मास्टर + Noun + SG + Masculine |
| मास्टरनी | | मास्टर(नी) + Noun |

| सोनार सोनारिन | Gender Inflection | सोनार + Noun + SG + Masculine सोनार(इन) + Noun + SG + Feminine |

### 4.2 Experiment 2

634 identified adjectives were tested and verified manually, among which 593 were analyzed correctly. Table 3 represents the corresponding output and inflection type for Maithili adjective.

**Table 4.** Result for Adjective Inflections.

| Input Words | Inflection Type | Output |
|---|---|---|
| निकहा निकही | Gender Inflection | निक(हा) + Adj. + SG + Masculine निकही(ही) + Adj. + SG + Feminine |
| जरलाहा जरलाही | Gender Inflection | जरल(आहा) + Adj. + SG + Masculine जरल(आही) + Adj. + SG + Feminine |
| मोटका मोटकी | Gender Inflection | मोट(का) + Adj. + SG + Masculine मोट(की) + Adj. + SG + Feminine |

### 4.3 Experiment 3

After completion of third experiment the system has correctly analyzed most of the regular and irregular verb on 14,312 Maithili verb forms, which was verified manually. Few correct analyses are given in Table 4 with their inflection type.

**Table 5.** Result for Verb Inflection

| Input Words | Inflection Type | Output |
|---|---|---|
| अईलाह अइलीह | Gender Inflection | अईल(आह) + Verb + SG + Masculine अईल(ईह) + Verb + SG + Feminine |
| चलू चलह | Person Inflection | चल(उ) + Verb + Imperative + SG + Masculine चल(ह) + Verb + Imperative + SG + Masculine |
| जाई जो | Person Inflection | जा(ई) + Verb + Optative + SG + Masculine जा(ओ) + Verb + Optative + SG + Masculine |

### 4.4 Performance

Currently there is no such state of art for Maithili morphological analyzer therefore there is no question of comparison exist. We have executed our system against 500 words for each POS category chosen randomly from Maithili news portal '*esamaad*', and checked the accuracy of results. Table 5 shows the results, which indicates a high percentage of correct results varying from category to category.

**Table 6.** Overall performance of Maithili morphological analyzer

| Word Type Input | No. of Words | Percentage of Correct Results |
|---|---|---|
| Noun Inflections | 500 | 92 |
| Adjective Inflections | 500 | 93 |
| Verb Inflections | 500 | 95 |
| Pronouns Inflections | 500 | 98 |
| Adverb Inflections | 500 | 98 |

# 5   Conclusion

In this paper, we have described the Maithili morphological analyzer which handles the inflectional property of language. A finite state transducer based approach has used to model the system, where a set of rules govern the system to produce the inflection of words from different grammatical categories. A series of experiments were conducted to evaluate the correctness of the system and an overall performance measure to evaluated gross result for fixed number of words falling in each category. Work on derivational category for Maithili will be the topic of interest for future research which will lead to a wide variety of computational support for this language.